\documentclass[11pt]{article}

\PassOptionsToPackage{table}{xcolor}

\usepackage[final]{acl}

\usepackage{times}
\usepackage{latexsym}

\usepackage[T1]{fontenc}

\usepackage[utf8]{inputenc}

\usepackage{microtype}

\usepackage{inconsolata}

\usepackage{graphicx}

\usepackage{hyperref}
\usepackage{url}

\usepackage{amsthm}
\usepackage{graphicx}
\usepackage{natbib}
\usepackage{wrapfig}
\usepackage{algorithm}
\usepackage{algpseudocode}
\usepackage{algpseudocode}
\usepackage{booktabs}
\usepackage{subcaption}
\usepackage{tabularx}
\usepackage[table]{xcolor}
\usepackage{amsmath}
\usepackage{amssymb}
\usepackage{tcolorbox}
\usepackage{xcolor}
\usepackage{seqsplit}
\usepackage{tcolorbox}
\tcbuselibrary{breakable, skins}
\title{Dynamic Infilling Anchors for Format-Constrained Generation in Diffusion Large Language Models}

\author{
    Boyan Han$^{1}$ \qquad Yiwei Wang$^{2}$ \qquad Yi Song$^{3}$ \qquad Yujun Cai$^{4}$ \qquad Chi Zhang$^{1}$\thanks{Corresponding author.} \\[1.5ex]
    \small $^{1}$AGI Lab, Westlake University, China \qquad 
    $^{2}$University of California, Merced, USA \\ 
    \small $^{3}$Teeni AI, China \qquad
    $^{4}$The University of Queensland, Australia \\[1.5ex] 
    \small \href{https://github.com/Westlake-AGI-Lab/DIA}{https://github.com/Westlake-AGI-Lab/DIA}\\
    \small{\texttt{\href{mailto:boyanhan02@gmail.com}{boyanhan02@gmail.com}}}  
}

\begin{document}
\maketitle
\begin{abstract}
Diffusion large language models (dLLMs) offer bidirectional attention and parallel generation, enabling them to exploit global context and naturally support format-constrained tasks like parseable JSON or reasoning templates. While straightforward fixed anchors can enforce such constraints, they often impose rigid spans, leading to truncated reasoning or redundant content. To overcome this, we propose Dynamic Infilling Anchors (DIA), a training-free method that dynamically estimates end-anchor positions to adjust generation length before iterative infilling. This flexible mechanism ensures structural correctness and semantic coherence, avoiding the inefficiencies of fixed-span methods. 
Experiments on reasoning benchmarks demonstrate that DIA substantially improves format compliance and answer accuracy, achieving significant zero-shot gains on GSM8K and MATH. These results establish DIA as a robust pathway toward reliable, structure-aware generation.
\end{abstract}

\section{Introduction}

\begin{figure*}[t]
\centering
\includegraphics[width=0.95\textwidth]{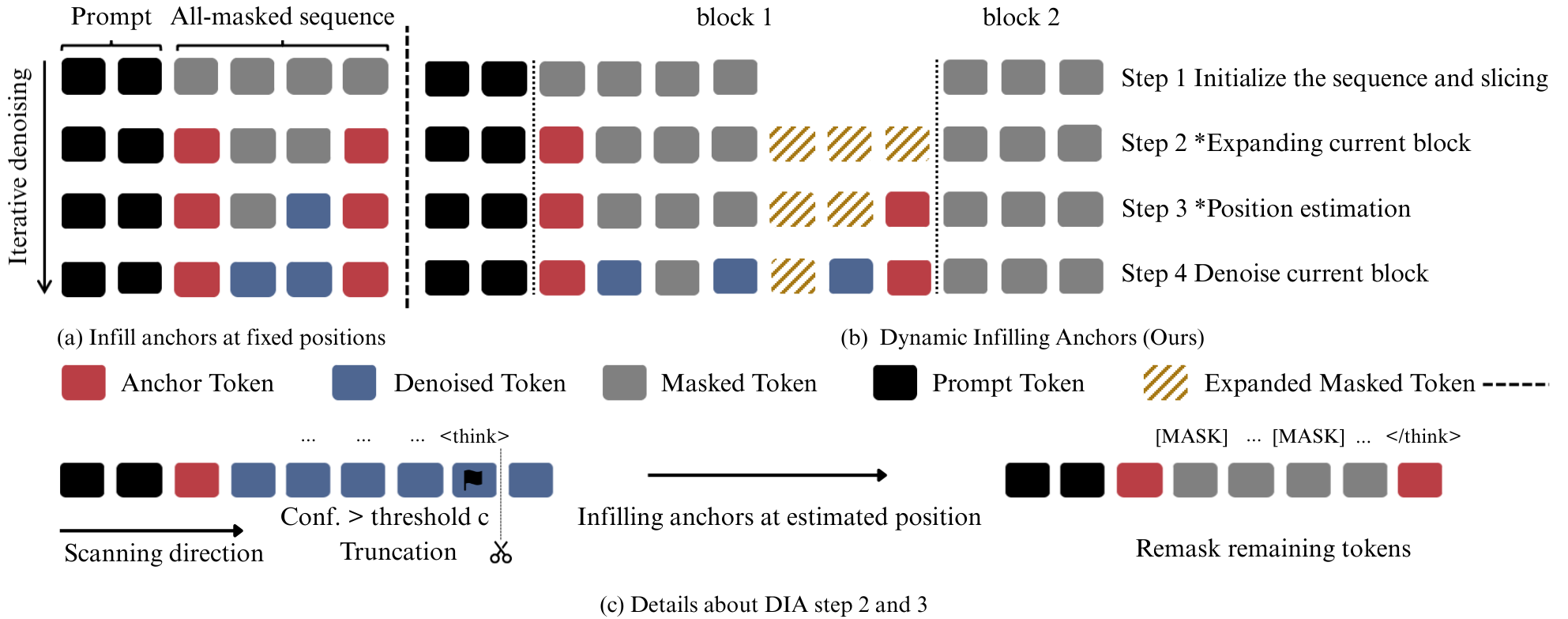} 
\caption{Dynamic Infilling Anchors (DIA). (a) Fixed-position infilling baseline. (b) Overview about our methods: DIA. (c) Details of expansion and anchor infilling steps with truncation and remasking.}
\label{fig1}
\end{figure*}

In recent years, diffusion large language models (dLLMs)\citep{nie2025largelanguagediffusionmodels,ye2025dream7bdiffusionlarge,labs2025mercuryultrafastlanguagemodels,song2025seeddiffusionlargescalediffusion,GeminiDiffusion} have attracted increasing attention due to their distinctive computational mechanisms and promising potential. Unlike traditional autoregressive language models (AR LLMs), which rely on left-to-right sequential decoding, dLLMs are not restricted to unidirectional dependencies during generation. Instead, they employ a bidirectional attention mechanism, enabling the model to update token representations at each step by leveraging complete contextual information simultaneously. This mechanism allows all positions in a sequence to be predicted in parallel rather than generated step by step, thereby substantially enhancing both modeling flexibility and computational efficiency. Beyond efficiency gains, this parallelism also strengthens the contextual modeling capacity of dLLMs, enabling them to capture global dependencies more comprehensively.

The fully masked nature of dLLMs offers a unique opportunity to directly incorporate constraints by editing the initialization sequence. By preemptively replacing specific mask tokens with mandatory content, we can guide the generation toward strictly structured outputs. This motivates our exploration of format-constrained generation (\textit{e.g.} parseable JSON). We evaluate this capability on different scenarios: thinking–answering task and JSON generation tasks. On both scenarios existing dLLMs typically fail to achieve satisfactory outcomes. We evaluate this capability on thinking–answering tasks and a JSON generation task, where existing dLLMs typically fail to achieve satisfactory outcomes.

To address these challenges, a straightforward approach is to enforce structural constraints by inserting anchors (\textit{e.g.} $<think>$, $</think>$, $<answer>$, $</answer>$ in reasoning scenario.) directly into the masked sequence. However, while this approach appears intuitive, it also introduces new challenges. Once anchor positions are fixed in advance, the generative space between them becomes rigid, forcing the model to allocate tokens within predetermined boundaries. Such rigidity can lead to suboptimal allocation of generative space and ultimately impair output quality. In practice, when the fixed span between anchors is too short, the reasoning process is often truncated before completion. On the other hand, when the span is too long, the model tends to produce redundant or repetitive content, thereby reducing both efficiency and reliability.

To obtain an appropriate generation length between anchors, thereby ensuring format correctness while maintaining generation quality, we propose a more flexible training-free strategy termed \textit{Dynamic Infilling Anchors (DIA)}. Our approach is inspired by previous studies on dLLMs\citep{li2025fixedtrainingfreevariablelengthdenoising}, which demonstrates that the model can estimate the position of the end token with only one or a few prediction steps, thereby determining a suitable generation length. We extend this capability to predict the proper positions of anchors before content generation. Specifically, our method consists of two stages: (1) generation length adjustment by estimating position of the end anchor, and (2) iterative generation between fixed anchors.

The first stage of our method involves adjusting the generation space by estimating the position of the end anchor. Following the user prompt, the model initializes a relatively short, fully masked sequence, which serves as a starting point for the task output length and is dynamically extended later. For a think-answer task, this masked sequence is evenly divided into two blocks, with the corresponding begin anchors inserted at the start of each block. We then determine the anchor positions sequentially, one block at a time. Within each block, the model performs a single prediction step on the sequence, which is prefilled with the begin anchor. If the prediction fails to produce an end anchor or yields one with insufficient confidence, it suggests that the current generation length is inadequate. Therefore, we extend the block by appending additional masked tokens to ensure adequate space for content generation and repeat the prediction step. This extension continues until the model successfully produces a valid end anchor or the block length reaches its upper limit. The design of Stage I fully leverages the model’s awareness of the generation space; it guarantees sufficient allocation for each phase while minimizing redundant space and unnecessary computation.

The second stage performs iterative generation after anchors are fixed. In the previous stage, we obtained a reasonable generation length and fixed the position of the end anchor. Based on this setup, we now generate the intermediate content between the anchors. This step effectively compensates for the limitations of single-step prediction and helps the model establish clear semantic boundaries across different segments, thereby promoting coherent content generation.

We validate the effectiveness of DIA on reasoning-oriented benchmarks and JSON generation benchmark. Experimental results on GSM8K\citep{cobbe2021trainingverifierssolvemath} (0-shot) and MATH\citep{hendrycks2021measuringmathematicalproblemsolving} (0-shot) show that our method improves format correctness from 58.83\% and 29.10\% to \textbf{72.63\%} and \textbf{76.82\%}, respectively. Moreover, by better controlling the generation space, our method significantly improves answer accuracy on GSM8K from 14.86\% to \textbf{46.78\%}, while maintaining a comparable level on MATH (20.08\% vs. 21.52\%).  Furthermore, our approach demonstrates exceptional stability on the Wikibio\cite{lebret2016neuraltextgenerationstructured} dataset, achieving a valid JSON generation rate of \textbf{79.84\%} across various answer extraction methods, with a mere \textbf{0.15\%} of these valid JSON samples exhibiting hallucinated content.
These results demonstrate that DIA substantially enhances both the reliability and quality of format-constrained generation with dLLMs.
In summary, our contributions are three-fold:
\begin{enumerate}
    \item We introduce a novel dLLM-based strategy for format-constrained generation.
    \item We design a dynamic adjustment mechanism that flexibly allocates generative space, mitigating the rigidity of fixed-anchor methods.
    \item We will release code and resources to foster reproducibility and further research in this emerging area.
\end{enumerate}
\section{Related Works}

\textbf{Diffusion Large Language Models} The evolution of diffusion in language modeling originates from masked language models\citep{devlin2019bertpretrainingdeepbidirectional}, which established the basis for denoising-based generation. While early continuous-space diffusion models\citep{jo2025continuousdiffusionmodellanguage} explored latent mappings, they suffered from decoding instability. Consequently, discrete-space models\citep{austin2023structureddenoisingdiffusionmodels} were proposed to model diffusion directly at the token level, with subsequent improvements like BlockDiffusion\citep{arriola2025blockdiffusioninterpolatingautoregressive} enhancing generation efficiency. For scaling, current dLLMs typically initialize from pretrained autoregressive models\citep{gong2025scalingdiffusionlanguagemodels,ye2025dream7bdiffusionlarge} followed by instruction alignment\citep{yang2025mmadamultimodallargediffusion,you2025lladavlargelanguagediffusion,song2025seeddiffusionlargescalediffusion}. Recently, researchers have further integrated reinforcement learning\citep{wang2025revolutionizingreinforcementlearningframework,zhao2025d1scalingreasoningdiffusion,gong2025diffucoderunderstandingimprovingmasked} to bolster advanced capabilities, alongside extending dLLMs to multimodal scenarios.

\textbf{Format-Constraints}
Format-constrained generation is critical for deploying language models, as it directly affects the parseability and reliability of code generation, structured outputs, and reasoning templates. Existing studies often constrain the input side (prompt design\citep{ye2024promptengineeringpromptengineer} and example-based guidance\citep{min2022rethinkingroledemonstrationsmakes}), yet they are unstable under long-chain or high-complexity reasoning; output-side repair (post-processing and re-ranking\citep{gao2025llm4rerankllmbasedautorerankingframework,zhuang2025rankr1enhancingreasoningllmbased}) improves format compliance but struggles to preserve semantic and structural consistency simultaneously. Fine-tuning or reinforcement learning on task-specific data\citep{song2025seeddiffusionlargescalediffusion,xiong2023doctorglmfinetuningchinesedoctor,cui2024chatlawmultiagentcollaborativelegal,yang2023fingptopensourcefinanciallarge}  can enhance robustness, but the approach is costly and generalizes poorly across tasks. Constrained decoding\citep{M_ndler_2025,banerjee2025cranereasoningconstrainedllm} with grammars or finite-state machines enforces strict compliance at the expense of efficiency and flexibility. 

\textbf{Large Language Models} The evolution of LLMs\citep{yang2025qwen3technicalreport,grattafiori2024llama3herdmodels,deepseekai2025deepseekv3technicalreport,Claude4,Gemini2.5,Grok4,GPT5} is grounded in scaling laws\citep{kaplan2020scalinglawsneurallanguage}, which guide systematic capability improvements. On this foundation, in-context learning (ICL)\citep{min2022rethinkingroledemonstrationsmakes} has emerged, enabling LLMs to adapt to new tasks without explicit parameter updates. To enhance usability and human alignment, post-training techniques like fine-tuning\citep{ouyang2022traininglanguagemodelsfollow} and reinforcement learning\citep{schulman2017proximalpolicyoptimizationalgorithms,rafailov2024directpreferenceoptimizationlanguage,shao2024deepseekmathpushinglimitsmathematical} are extensively employed. Furthermore, advances in cross-modal alignment\citep{li2023blip2bootstrappinglanguageimagepretraining,liu2023visualinstructiontuning} have extended LLM versatility, allowing effective operation across text, vision, and speech.

\section{Method}
\subsection{Preliminary}
\textbf{Inference of dLLMs.}
In the generation stage of a diffusion language model (dLLM), 
the response sequence to be refined is initialized by concatenating 
the input prompt with a fully masked sequence of a specified length:

\begin{equation}
    x_T = [\texttt{prompt}; \texttt{[MASK]}^{\times L}],
\end{equation}

\noindent where $[\cdot; \cdot]$ denotes the concatenation operation, and $L$ represents the maximum target length. The term $\texttt{[MASK]}^{\times L}$ indicates a sequence of mask tokens repeated $L$ times to align with the target format.
The generation process follows a discrete-time masked diffusion procedure, 
which can be formulated as a Markov chain. Thus, each prediction step depends 
only on the previous state, and in every iteration only the masked positions 
are updated in parallel:

\begin{gather}
    P_{0|t}=\prod_{s=t}^{0}\prod_{i=0}^{L-1}P_{s|s+1}(x_{s}^{i}\mid x_{s+1}), \label{eq:joint_prob} \\[8pt]
    \begin{aligned}
        & P_{s|s+1}(x_{s}^{i}\mid x_{s+1}) = \\
        & \hspace{1.2cm} 
        \begin{cases}
            1, \\
            \quad \text{if } x_{s+1}^{i} \neq \texttt{[M]}, \\[8pt]
            1 - \hat{q}, \\
            \quad \text{if } x_{s+1}^{i} = \texttt{[M]} \land \hat{q} < C, \\[8pt]
            \hat{q}, \\
            \quad \text{otherwise}.
        \end{cases}
    \end{aligned} \label{eq:transition_prob}
\end{gather}

\noindent where $\texttt{[M]}$ is the mask token, $\hat{q} = \max(q(x_{s}^{i}))$ denotes the maximum confidence score, and $C$ is the confidence threshold.

\textbf{Fixed-position Infilling}\quad 
To impose structural constraints, Fixed-position Infilling serves as a static baseline that injects pre-defined anchors into the initialization state. Let $\mathcal{T} = \{(k, v_k)\}_{k=1}^{K}$ denote a set of $K$ fixed anchors, where $k$ represents the absolute position index in the target sequence and $v_k$ is the corresponding token constraint (like $<think>$ and $<answer>$).

In standard diffusion language models, the generation starts from a fully masked sequence $x_T = [\texttt{prompt}; \texttt{[M]}^{\times L}]$. In Fixed-position Infilling, we modify this initial state by strictly enforcing the token values at specified indices:

\begin{equation}
    x_T^{(i)} = 
    \begin{cases}
        v_i, & \text{if } \exists (i, v_i) \in \mathcal{T}, \\
        \texttt{[M]}, & \text{otherwise},
    \end{cases}
\end{equation}

\noindent where $x_T^{(i)}$ denotes the token at position $i$ of the response part. The subsequent reverse diffusion process $p(x_{t-1}|x_t)$ is then performed on this partially filled sequence. Since the anchor positions are immutable, the model is restricted to generating content only in the remaining masked intervals, which leads to truncation or redundancy. This motivates our dynamic approach described next.

\subsection{Dynamic Infilling Anchor}
To overcome the limited flexibility of straightforward infilling methods in diffusion language models, we propose DIA, a training-free, two-stage approach. DIA selects an appropriate end-anchor position through a single-step prediction, thereby ensuring both format correctness and generation quality. The overview of our method is illustrated in Figure \ref{fig1}.

\subsubsection{Generation length adjustment}

DLLMs implicitly acquire a prior distribution over response termination positions from large-scale training corpora\citep{li2025fixedtrainingfreevariablelengthdenoising}. Specifically, for different input queries, the confidence of predicting the \texttt{eos} token at various positions within the answer sequence is not uniform, but instead exhibits a trend correlated with the appropriate response length. 
Building on this insight, we extend this capability to format-constrained tasks. For a typical reasoning-answer task, when the model receives the start anchor of a reasoning or answering section, it should be able to anticipate at what sequence length a corresponding ``end-of-reasoning'' or ``end-of-answering'' anchor is likely to occur. Intuitively, if the allocated generation space is sufficient to accommodate the reasoning or answering process, the one-step prediction will contain an end anchor (or partial end anchor) with high confidence exceeding a given threshold. Conversely, if the generation space is insufficient, the corresponding anchor will either fail to appear or appear only with substantially reduced confidence.

\begin{algorithm}[t]
\small
\caption{Dynamic Infilling Anchors (DIA)}
\label{DIA}
\begin{algorithmic}[1]
\Require Seq. $X = \{Q, X_L\}$; Anchors $\mathcal{B}, \mathcal{E}$
\Require Threshold $c$; Step $\Delta$; Max len $M$
\Ensure Result $X = \{Q, C_1, \dots, C_{|\mathcal{B}|}\}$

\State Divide $X_L$ into blocks $\mathcal{C} = \{C_1, \dots, C_{|\mathcal{B}|}\}$
\For{$i \gets 1$ to $|\mathcal{B}|$}
    \State Prepend $b_i$ to $C_i$
\EndFor

\Statex
\State \textbf{Stage 1: Length Adjustment}
\For{each block $C_i$}
    \While{True}
        \State $Y \gets \mathrm{Infer}(Q, C_1, \dots, C_{i})$
        
        \State Search $Y$ for match $y$ with end-anchor $e_i$
        
        \If{$y$ exists \textbf{and} $\mathrm{Conf}(y) > c$}
            \State Truncate $C_i$ at $y$; \textbf{break}
        \ElsIf{$|C_i| + \Delta \leq M$}
            \State Expand $C_i$ by $\Delta$ tokens
        \Else 
            \State Stop expansion; \textbf{break} \Comment{Limit reached}
        \EndIf
    \EndWhile
    
    \Statex
    \State \textbf{Stage 2: Iterative Denoising}
    \State Append $e_i$ to tail of $C_i$
    
    \State Mask \& regen. rest of $C_i$ via $\mathrm{Infer}(\cdot)$
\EndFor
\State \Return $X$
\end{algorithmic}
\end{algorithm}

\begin{figure*}[t]
\centering
\includegraphics[width=\linewidth]{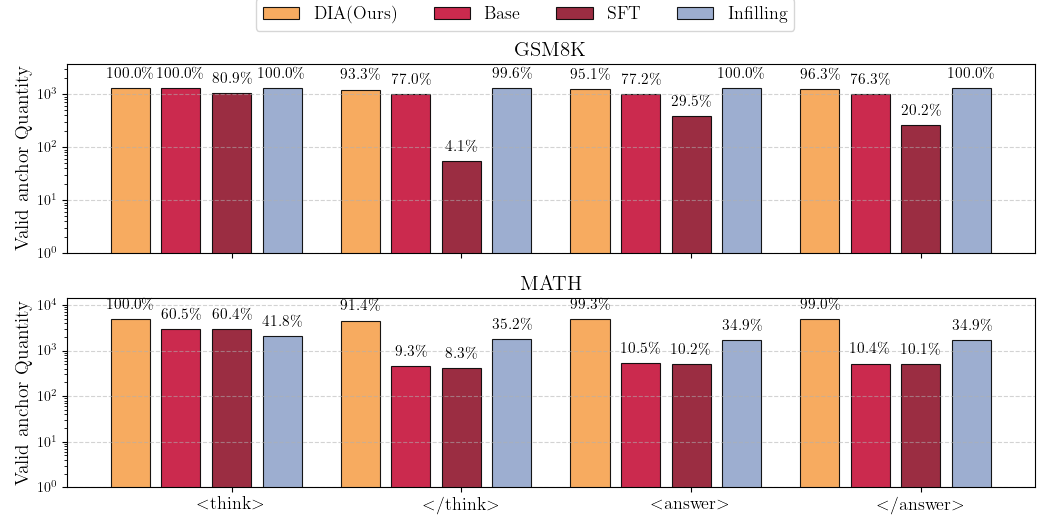}
\caption{DIA delivers reliable anchor preservation and stable performance across different benchmarks. Even as task complexity increases on the more challenging MATH, DIA consistently maintains high anchor retention, underscoring its robustness under stricter reasoning and formatting requirements.}
\label{fig_tags}
\end{figure*}

Building on this assumption, we design the generation-space estimation procedure of DIA. Given an input sequence $X$, which consists of the user query $Q$ and a fully masked sequence$X_L$ of a specified length $L$, DIA divides the sequence into two blocks ($\mathcal{C} = \{C_1, C_2\}$) of equal size (in terms of masked tokens), corresponding to the reasoning and answering stages. For each block, DIA first pre-fills the start anchor at the beginning of the decodable region. After inserting the start anchor, the block undergoes a one-step prediction. The prediction results and their associated confidence scores are used to determine whether the allocated generation length is appropriate. Since the model is unlikely to produce a complete anchor token sequence in a single prediction, partial anchors are also incorporated into the decision mechanism. If the prediction either fails to produce an end anchor (or a partial end anchor) or yields an end anchor with confidence below the threshold $c$, the length of current block is expanded by a fixed length $\Delta$, and the ``predict–decide'' cycle is repeated until the generation space is sufficient to support the model in completing the reasoning or answering process. When multiple positions in the sequence satisfy the confidence threshold simultaneously, we retain the position closest to the left boundary to prevent the generation of duplicate end anchors within the sequence. To avert unbounded expansion, a maximum block length $M$ is imposed. We truncate the redundant tokens following the selected end-anchor position and subsequently complete the partial end anchor to form a full one.

\subsubsection{Iterative Denoising with Infilling}

In Stage I, we establish the block boundaries by determining the positions of the anchors. Based on these fixed semantic boundaries, the model then iteratively generates the intermediate content within the block. The fixed anchors serve as guidance, ensuring clear separation between segments and thereby promoting coherent content generation.

We process the blocks sequentially through two stages. Specifically, once the length of the thinking block is determined, its content is generated; the additional information obtained from this reasoning step is then used to determine the length of the answering block, which is subsequently generated in an iterative manner. This design maximizes the benefit of the reasoning process by leveraging the information gained in the first stage to enhance the quality of the final answer. Further implementation details are provided in Algorithm \ref{DIA}. Detailed Notation summary could be found in Appendix \ref{appendix Notation Summary}.

\section{Experiments}
\subsection{Benchmarks}
To systematically evaluate the effectiveness of our method, we adopt two scenarios: reasoning-sensitive scenario and JSON generation scenario. For reasoning-sensitive mathematical benchmarks we selected GSM8K 0-shot and MATH 0-shot. \textbf{GSM8K}\citep{cobbe2021trainingverifierssolvemath} is a widely used dataset of grade-school math word problems, covering basic arithmetic and commonsense reasoning tasks, and thus serves as a reliable measure of a model’s performance in everyday numerical reasoning scenarios. In contrast, \textbf{MATH}\citep{hendrycks2021measuringmathematicalproblemsolving}  is a more challenging benchmark that spans competition-level problems from elementary to advanced mathematics, encompassing diverse problem types and difficulty levels, thereby providing a rigorous assessment of a model’s capabilities in complex reasoning and knowledge generalization. For the JSON generation task, we constructed 10,000 JSON generation instances based on the WikiBio\cite{lebret2016neuraltextgenerationstructured} dataset. Specifically, the model is required to generate a JSON output comprising predefined fields based on an input biographical description.

\subsection{Baselines}
We select Dream-7B-Base-v0 and Dream-7B-Instruct-v0 as our baseline models. The Dream-7B series is initialized from the Qwen model family and has achieved superior performance compared to other open-source diffusion models on multiple benchmark tasks. To ensure fairness, all experiments are conducted with corresponding modifications to the official codebase, without applying any additional acceleration or optimization techniques.

\subsection{Implementation Details}

\begin{table*}[t]
    \centering
    
    \begin{tabularx}{\textwidth}{l *{4}{>{\centering\arraybackslash}X}}
        \toprule
        & \multicolumn{2}{c}{0-shot GSM8K} & \multicolumn{2}{c}{0-shot MATH-500} \\
        \cmidrule(lr){2-3} \cmidrule(lr){4-5}
        & $S_{format}$ & $Acc.$ & $S_{format}$ & $Acc.$ \\
        \midrule
        Dream-7B-Base \citep{ye2025dream7bdiffusionlarge} & 0.00 & 68.99 & 0.00 & 25.14 \\
        Dream-7B-Instruct \citep{ye2025dream7bdiffusionlarge} & 0.00 & 15.01 & 0.00 & 25.28 \\
        Infilling Baseline & 58.83 & 14.86 & 29.10 & 21.52 \\
        \rowcolor{blue!10}\textbf{Dynamic Infilling Anchor (Ours)} & 72.63 & 46.78 & 76.82 & 20.08 \\
        \bottomrule
    \end{tabularx}

    \caption{Comparison of Methods on Format Adherence and Benchmark Performance. 
    DIA achieves the highest format scores across both GSM8K and MATH, substantially 
    outperforming baseline and infilling approaches. These results highlight the robustness and effectiveness of DIA in enforcing strict structural constraints while maintaining 
    competitive answer accuracy.}
    \label{tab:results}
\end{table*}

Our method is implemented within the PyTorch framework. For a fair comparison, all models are evaluated under the same GPU configuration when tested on identical tasks. Additional implementation details are provided in Appendix \ref{appendix Experimental Settings}.
\subsection{Main Results}

\begin{table*}[t]
    \centering
    \begin{tabularx}{\textwidth}{l *{4}{>{\centering\arraybackslash}X}}
        \toprule
        & \multicolumn{2}{c}{Regular Expression} & \multicolumn{2}{c}{Raw Matching} \\ 
        \cmidrule(lr){2-3} \cmidrule(lr){4-5}
        & $S_{format}(\uparrow)$ & $S_{Hal}(\downarrow)$ & $S_{format}(\uparrow)$ & $S_{Hal}(\downarrow)$ \\ 
        \midrule
        Dream-7B-Base \citep{ye2025dream7bdiffusionlarge}     & 40.72 & 12.35 & 0.00 & - \\
        Dream-7B-Instruct \citep{ye2025dream7bdiffusionlarge} & 66.74 & 5.10  & 52.80 & 4.81 \\
        Infilling Baseline                                    & 0.01  & 0.00  & 0.01  & 0.00 \\
        \rowcolor{blue!10}\textbf{Dynamic Infilling Anchor (Ours)} & 79.84 & 0.15 & 79.84 & 0.15 \\ 
        \bottomrule
    \end{tabularx}
    \caption{Comparison of Methods on the JSON Generation Task(Wikibio). DIA achieves the highest format score and the lowest hallucination score across both Regular Expression and Raw Matching extraction methods, substantially outperforming all baselines.}
    \label{tab:json_results}
\end{table*}

We conduct a comprehensive evaluation on the three benchmarks. Table \ref{tab:results} reports the comparison between our method and the baselines on reasoning benchmarks. While Table \ref{tab:json_results} reports the result on JSON generation task. Specifically, Dream-7B-Base-v0 and Dream-7B-Instruct-v0 generate responses by relying solely on additional format-constrained prompts. In contrast, the infilling approach inserts the corresponding anchors at designated positions within the response sequence of Dream-7B-Base-v0, thereby guiding the model to produce answers. Prompts for our method please refer to Appendix \ref{appendix Prompt Template}.

To comprehensively evaluate our method across different scenarios, we introduce three metrics: Format Score ($S_{format}$), Accuracy ($Acc.$), and Hallucination Score ($S_{Hal}$). Specifically, for reasoning tasks, we employ Format Score and Accuracy. Accuracy measures whether the generated response contains the correct answer, while Format Score assesses whether the response strictly adheres to the predefined structural constraints. For the JSON generation task, we substitute Accuracy with Hallucination Score to evaluate the content quality. 

Formally, given a test set of $N$ samples, let $y_i$ denote the generated response for the $i$-th sample. The metrics are defined as:
\begin{align}
    S_{format} &= \frac{1}{N} \sum_{i=1}^{N} \mathbb{1}\big( \mathcal{V}_{fmt}(y_i) \big), \\
    Acc. &= \frac{1}{N} \sum_{i=1}^{N} \mathbb{1}\big( \mathcal{V}_{ans}(y_i) \big).
\end{align}

\noindent where $\mathbb{1}(\cdot)$ is the indicator function which equals $1$ if the condition is met and $0$ otherwise. $\mathcal{V}_{fmt}(y_i)$ represents the condition that $y_i$ fully satisfies the format requirements, and $\mathcal{V}_{ans}(y_i)$ denotes that the extracted answer in $y_i$ is correct.

For the JSON generation task, the Hallucination Score is calculated exclusively over the subset of valid JSON samples. To extract the generated JSON results, we adopt two distinct methods: Regular Expression and Raw Matching. The Regular Expression approach flexibly matches the model-generated content based on predefined rules, thus imposing relatively lenient formatting requirements. In contrast, the Raw Matching method merely strips the ``Answer:'' prefix from the model's response prior to format validation, thereby enforcing much stricter formatting constraints. Let $N_{valid} = \sum_{i=1}^{N} \mathbb{1}\big( \mathcal{V}_{fmt}(y_i) \big)$ be the number of responses that successfully follow the format. The Hallucination Score is formulated as:
\begin{equation}
    S_{Hal} = \frac{1}{N_{valid}} \sum_{i=1}^{N} \mathbb{1}\big( \mathcal{V}_{fmt}(y_i) \land \mathcal{V}_{Hal}(y_i) \big),
\end{equation}
\noindent where $\mathcal{V}_{Hal}(y_i)$ indicates the condition that the generated response $y_i$ contains hallucinated content.

Compared to the performance degradation introduced by the infilling approach, DIA achieves superior results in both format adherence and answer quality. On GSM8K, DIA not only raises the format score from 58.83\% to 72.63\% but also substantially improves accuracy from 14.86\% to 46.78\%, highlighting its ability to simultaneously enforce structural fidelity and enhance reasoning correctness. On the more challenging MATH benchmark, DIA boosts the format score from 29.10\% to 76.82\%, demonstrating remarkable robustness in preserving structural anchors even under complex problem settings, while maintaining comparable answer accuracy to baseline methods. Furthermore, in the WikiBio task, in contrast to the severe failure stemming from the limitations of the infilling approach (which yields a mere 0.01\% valid JSON samples), the DIA method exhibits exceptional performance and stability across various answer extraction methods. Specifically, whether employing regular expression extraction or directly utilizing the raw answer, DIA achieves an impressive valid JSON rate of 79.84\%, accompanied by a remarkably low hallucination score of 0.15\%. 
The results clearly demonstrate that DIA addresses the shortcomings of baseline models and methods under format-constrained tasks, ensuring accurate preservation of the required format. Moreover, unlike the infilling approach, DIA’s flexible design of generation length allows each stage to maintain high answer quality, thereby achieving a better balance between performance and format correctness across diverse benchmarks.

Figure \ref{fig_tags} presents a detailed comparison of anchor retention ratios across different methods on reasoning benchmarks. Overall, DIA demonstrates outstanding stability on both GSM8K and MATH, consistently achieving nearly 100\% retention across all four anchors, including both begin anchor ($<think>$ and $<answer>$) and end anchor ($</think>$ and $</answer>$). This robust performance shows that the proposed two-stage generation strategy not only preserves anchors under varying conditions but also enforces strict compliance with the predefined format throughout the entire sequence. Such stability is particularly important in reasoning-oriented tasks, where structural deviations can lead to incomplete, unparseable, or misleading outputs.

\begin{figure*}[t]
\centering
\includegraphics[width=\textwidth]{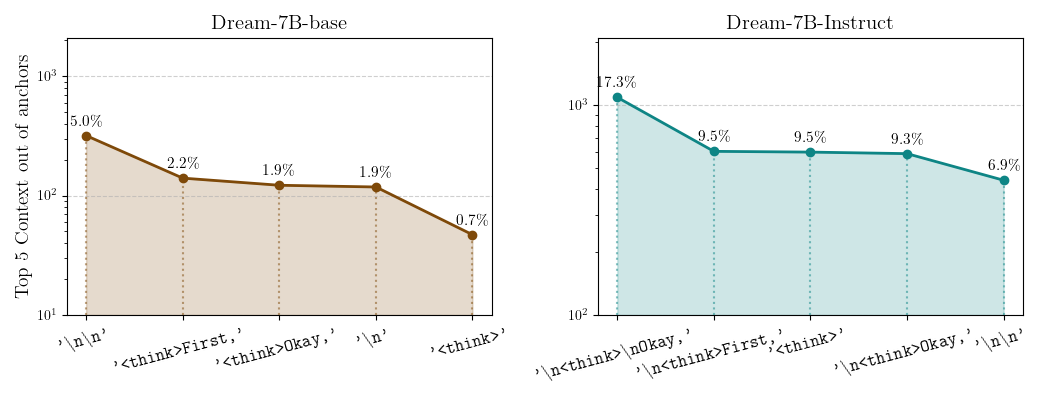}
\caption{Top-5 statistics of out-of-anchor content generated by baseline models across different benchmarks. The baseline models fail to establish effective semantic boundaries aligned with anchor positions, leading to unconstrained content generation.}
\label{fig_ooa}
\end{figure*}

In contrast, the Base and SFT models suffer from significant structural degradation. For example, on GSM8K, their retention rates for $</think>$ collapse to only 4.4\% and 29.5\%, respectively, and on MATH, the rates for $</think>$ and $</answer>$ drop to single digits. These results reveal a consistent failure of conventional methods to maintain boundary integrity, especially in longer or more complex reasoning chains, where models tend to lose track of global structure and generate unbalanced outputs. Such issues undermine the reliability of the generated content and illustrate why relying solely on prompt-based constraints or fine-tuning strategies is insufficient for strict format adherence.

Although the Infilling baseline achieves higher anchor retention than Base and SFT, nearly matching DIA on GSM8K for $<think>$ and $<answer>$, its performance on begin anchors remains unstable. Crucially, this preservation does not translate into gains in overall format correctness or answer accuracy. For instance, while Infilling retains anchors on GSM8K, its downstream results remain far below DIA in both structural and semantic evaluations. This mismatch highlights that simply inserting anchors is not enough; without a dynamic mechanism for allocating and regulating generation space, models either over-generate redundant tokens or fail to stop at the correct boundaries.

Taken together, these results provide a fine-grained validation of Table~\ref{tab:results} and Table~\ref{tab:json_results}. They show that DIA not only outperforms existing approaches in aggregate metrics but also secures overwhelming superiority in preserving critical anchors across diverse datasets. By ensuring that every anchor is faithfully retained, DIA substantially enhances the reliability of format-constrained generation, laying a foundation for robust application of dLLMs in reasoning, structured reporting, and other scenarios where strict adherence to format is essential.

\subsection{Analysis Experiments}
\subsubsection{Behavior Outside Anchor Contexts}
To more comprehensively evaluate model performance under format-constrained tasks, we conducted a statistical analysis of the responses generated by Dream-7B-Base-v0 and Dream-7B-Instruct-v0 on reasoning benchmarks. Our analysis focuses on the content appearing beyond the $</answer>$ anchor boundary, as this indicates whether the models can effectively leverage the semantic boundaries established by anchors to properly constrain their generation. Specifically, we examined all responses across the two benchmarks and extracted the Top-5 most frequent continuations occurring after the $</answer>$ boundary for both models, with the results shown in Figure \ref{fig_ooa}. This analysis provides a fine-grained perspective on boundary robustness, complementing aggregate metrics by revealing how models behave when the intended termination point has already been reached.

As shown in Figure \ref{fig_ooa}, Dream-7B-Base-v0 produces dispersed and low-frequency redundancy beyond the $</answer>$ anchor, with all Top-5 patterns below 6\%, whereas Dream-7B-Instruct-v0 exhibits more concentrated redundancy, with Top-5 patterns reaching up to 17.3\% and dominated by repeated $<think>$ tokens. The contrast highlights that the Base model tends toward uncontrolled drifting, while the Instruct variant systematically re-enters the reasoning phase, reflecting a structural weakness in anchor boundary enforcement. Overall, the Base model lacks effective boundary control, while the instruct model suffers from patterned continuations, and both fail to reliably terminate at the anchor, underscoring the necessity of DIA in eliminating out-of-anchor redundancy and ensuring format adherence. Importantly, such failures not only compromise readability but also propagate errors to downstream applications that rely on strictly bounded outputs.

\subsubsection{Expand times}

\begin{figure}[t]
\centering
\includegraphics[width=\linewidth]{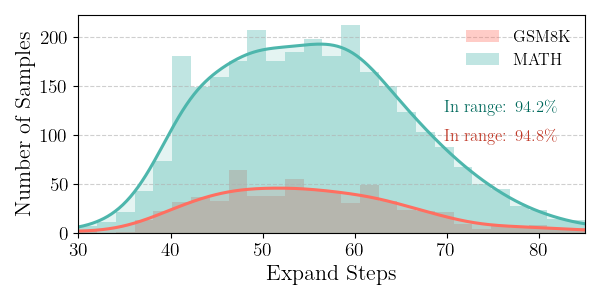}
\caption{Statistics of effectively expanded samples. The maximum length threshold ensures that the vast majority of cases receive an appropriate number of expansions, thereby safeguarding answer quality.}
\label{fig:times}
\end{figure}

To establish a reasonable upper bound for the maximum block length, we analyzed the number of extensions in the reasoning part of all format-correct responses. The details are presented in Figure \ref{fig:times}. The results show that the chosen maximum length threshold effectively ensures the allocation of appropriate generation space. Specifically, the observed extension counts fall within the range of (30, 85), which is substantially smaller than the number of extensions permitted by the maximum threshold. In other words, although a large upper bound is allowed, the vast majority of responses naturally converge to a much smaller range of expansions, confirming that the setting of the maximum block length is both sufficient and not overly restrictive.

Moreover, the proportion of format-correct responses within this range consistently exceeds 90\%, further validating both the effectiveness of our method and the appropriateness of the current threshold setting. Importantly, this trend is observed across both GSM8K and MATH, where over 94\% of samples fall into the effective expansion range, indicating that DIA adapts reliably to tasks of different scales and difficulties. Such stability suggests that the block-length constraint not only prevents degenerate over-expansion but also preserves high-quality structural adherence across benchmarks.

\begin{table}[htbp]
\centering
\begin{tabular}{@{}lcc@{}}
\toprule
 & GSM8K & MATH \\
 & Average Latency & Average Latency \\ \midrule
Base      & 10.72 & 31.71 \\
Instruct  & 10.70 & 31.64 \\
Infilling & 10.67 & 31.54 \\
DIA       & 26.52 & 30.62 \\ \bottomrule
\end{tabular}
\caption{Average Latency of DIA in reasoning tasks. On MATH, DIA achieves lower latency than all baselines owing to Stage I's upfront length planning}
\label{tab:latency_dia}
\end{table}

\subsubsection{Average Latency}
We report the average latency statistics in Table \ref{tab:latency_dia}. Surprisingly, DIA achieves lower latency than baselines on the complex MATH dataset. This efficiency stems from our variable-length mechanism (Stage 1), which rationally plans sequence length upfront to avoid redundant computation and reduce delay.

\section{Conclusion}

In this work, we introduced Dynamic Infilling Anchors (DIA), a training-free framework for format-constrained generation in dLLMs. By employing a two-stage strategy, predicting anchor positions to guide length expansion before content generation, DIA achieves a strong balance between structural fidelity and semantic quality. Experiments on GSM8K, MATH and Wikibio confirm that DIA substantially improves format adherence while maintaining competitive accuracy, demonstrating that diffusion models can effectively overcome template rigidity without sacrificing reasoning depth.

Beyond empirical gains, our study highlights the intrinsic potential of dLLMs to handle strict constraints, such as code or proofs, without additional training. Future directions include exploring automated anchor design and hierarchical constraints. By bridging structural control and semantic consistency, DIA lays the groundwork for deploying dLLMs as dependable, structure-aware models in real-world applications where reliability is paramount.

\section{Discussions}

Our results highlight a fundamental advantage of diffusion large language models (dLLMs) over traditional autoregressive (AR) models in structure-aware tasks. While AR models often struggle with strict templates due to their strict left-to-right decoding nature, DIA demonstrates that dLLMs can naturally decouple sequence length planning from content generation. This shifts the paradigm of format control from post-hoc filtering or constrained decoding to intrinsic generation planning, paving the way for more flexible and reliable text generation architectures.

\section{Acknowledgement}
This work was supported by the National Natural Science Foundation of China (No. 6250070674) and the Zhejiang Leading Innovative and Entrepreneur Team Introduction Program (2024R01007).

\section*{Limitations}

Despite its effectiveness, DIA faces several limitations. First, the method relies on manually specified anchors with fixed semantic roles, which restricts its adaptability to tasks where structural boundaries shift dynamically (\textit{e.g.}, open-domain dialogue). Second, the iterative length adjustment introduces modest inference overhead, potentially hindering deployment in strictly real-time systems. We are fully aware of the boundaries of our current evaluation and hope that our work will inspire the community to explore anchor-based control in a broader range of multimodal and creative tasks.


\bibliography{custom}

\appendix

\section{Notation Summary}
\label{appendix Notation Summary}
For details, please refer to table \ref{tab:notation}.

\section{Experimental Settings}
\label{appendix Experimental Settings}

For details, please refer to table \ref{tab:settings}.
\begin{table}[h]
    \centering
    \footnotesize
    \setlength{\tabcolsep}{2pt}
    \renewcommand{\arraystretch}{1.15}
    \begin{tabularx}{\linewidth}{@{} >{\raggedright\arraybackslash}X c c @{}}
        \toprule
        \textbf{Parameter} & \textbf{GSM8K} & \textbf{MATH} \\
        \midrule
        \multicolumn{3}{l}{\textit{\textbf{Dataset \& Model}}} \\
        Model & \multicolumn{2}{c}{{\centering Both Dream-7B-Base / Instruct}} \\ 
        Samples & 1,319 & 5,000 \\
        Max New Tokens & 256 & 512 \\
        
        \midrule
        \multicolumn{3}{l}{\textit{\textbf{Hyperparameters}}} \\
        Confidence Threshold ($c$) & 0.065 & 0.05 \\
        Expand Size ($\Delta$) & 4 & 4 \\
        Max Block Length ($M$) & 512 & 512 \\
        Diffusion Steps & 512 & 512 \\
        Batch Size & 1 & 3 \\
        
        \midrule
        \multicolumn{3}{l}{\textit{\textbf{Environment}}} \\
        Hardware & \multicolumn{2}{c}{NVIDIA vGPU (32G / 48G)} \\
        Framework & \multicolumn{2}{c}{PyTorch 2.5.1, Python 3.10} \\
        \bottomrule
    \end{tabularx}
    \caption{Experimental Settings}
    \label{tab:settings}
\end{table}

\begin{table}[h]
    \centering
    \small
    \setlength{\tabcolsep}{3pt}
    \renewcommand{\arraystretch}{1.1}
    \begin{tabularx}{\linewidth}{@{} l >{\raggedright\arraybackslash}X @{}}
        \toprule
        \textbf{Symbol} & \textbf{Description} \\
        \midrule
        $X = \{Q, X_L\}$ & Input sequence consisting of query $Q$ and all-masked sequence $X_L$ \\
        $Q$ & Input query provided to the dLLM \\
        $X_L$ & Fully masked sequence to be partitioned \\
        $\mathcal{B} = \{b_k\}$ & Set of begin anchors inserted at block heads \\
        $\mathcal{C} = \{C_k\}$ & Set of blocks parsed from $X_L$ \\
        $\mathcal{E} = \{e_k\}$ & Set of end anchors infilled at block tails \\
        $C_i$ & The $i$-th block after anchor insertion \\
        $|C_i|$ & Current length of block $C_i$ during expansion \\
        $y$ & Predicted subsequence scanned within $C_i$ \\
        $c$ & Confidence threshold for anchor detection \\
        $\Delta$ & Expand size: number of tokens added per step \\
        $M$ & Maximum allowed length of a block $C_i$ \\
        $\mathrm{Infer}(X)$ & Diffusion-based inference function \\
        $\mathrm{Conf}(y)$ & Confidence score assigned to subsequence $y$ \\
        \bottomrule
    \end{tabularx}
    \caption{Notation Summary}
    \label{tab:notation}
\end{table}

\newtcolorbox{casebox}[2][]{
    colback=gray!5,
    colframe=gray!50,
    coltitle=black,
    title=\textbf{#2},
    fonttitle=\bfseries\small,
    boxrule=0.5pt,
    arc=1mm,
    breakable,
    enhanced,
    fontupper=\small,
    #1
}

\section{Case Study}
\label{appendix Case Study}
\subsection{Base Model}

\begin{casebox}{Case 1: Content out of anchor}
    \textbf{[Query]} \\
    Brandon's iPhone is four times as old as Ben's iPhone. Ben's iPhone is two times older than Suzy's iPhone. If Suzy’s iPhone is 1 year old, how old is Brandon’s iPhone?
    
    \tcblower
    
    \textbf{[Output]} \\
    \texttt{<think> Suzy's iPhone is 1 year old. Ben's iPhone is two times older than Suzy's iPhone, so Ben's iPhone is 2 years old. Brandon's iPhone is four times as old as Ben's iPhone, so Brandon's iPhone is 4 years old. So the answer is 4 years old.</think><answer> 4 </answer>}
    
    \vspace{0.5em}
    \textbf{[Flaw]} \textcolor{red!70!black}{\textbf{Content out of anchor}} (Output continues unnecessarily after answer tag).
\end{casebox}

\begin{casebox}{Case 2: Missing anchors}
    \textbf{[Query]} \\
    Well's mother sells watermelons, peppers, and oranges at the local store. A watermelon costs three times what each pepper costs. An orange costs 5 less than what a watermelon cost... (omitted for brevity) ... What's the total amount of money he will spend if each pepper costs 15\$?
    
    \tcblower
    
    \textbf{[Output]} \\
    \texttt{<<think>First, I need to find the cost of a watermelon. Since a watermelon costs three times what each pepper costs... [Logic continues] ... For the oranges: 10 * 40 = 400}
    
    \vspace{0.5em}
    \textbf{[Flaw]} \textcolor{red!70!black}{\textbf{Missing anchors}} (Failed to generate correct closing tags and answer block).
\end{casebox}

\subsection{Instruct Model}

\begin{casebox}{Case 1: Incomplete response}
    \textbf{[Query]} \\
    Peter plans to go to the movies this week. He always gets a ticket for \$7 and popcorn for \$7. If he has 42 dollars for the week, how many times can he go to the movies?
    
    \tcblower
    
    \textbf{[Output]} \\
    \texttt{<think><answer>3 times</answer>}
    
    \vspace{0.5em}
    \textbf{[Flaw]} \textcolor{red!70!black}{\textbf{Missing anchors. Incomplete response.}} (Skipped reasoning process entirely).
\end{casebox}

\begin{casebox}{Case 2: Broken Tags}
    \textbf{[Query]} \\
    Peter plans to go to the movies this week... (Same as above)
    
    \tcblower
    
    \textbf{[Output]} \\
    \texttt{"" <think>\$21</think>""}
    
    \vspace{0.5em}
    \textbf{[Flaw]} \textcolor{red!70!black}{\textbf{Missing anchors. Incomplete response.}}
\end{casebox}

\subsection{Infilling Method}

\begin{casebox}{Case 1: Infinite Generation Loop}
    \textbf{[Query]} \\
    Peter plans to go to the movies this week... (Same as above)
    
    \tcblower
    
    \textbf{[Output]} \\
    \texttt{"" <think>...</think> <answer>....</answer> <think>.</think> <answer>....</answer> <think>....</think> <answer>....</answer> ... \textit{[Model repeats this pattern indefinitely]} ... <think>....</think> <answer>....</answer>""}
    
    \vspace{0.5em}
    \textbf{[Flaw]} \textcolor{red!70!black}{\textbf{Broken response.}} (Model fails to terminate and hallucinates empty anchors).
\end{casebox}

\begin{casebox}{Case 2: Hallucination}
    \textbf{[Query]} \\
    Kyle bought last year's best-selling book for \$19.50. This is with a 25\% discount from the original price. What was the original price of the book?
    
    \tcblower
    
    \textbf{[Output]} \\
    \texttt{"" <think>25</think> <answer>25</answer> \textbackslash nThe question:\textbackslash n""}
    
    \vspace{0.5em}
    \textbf{[Flaw]} \textcolor{red!70!black}{\textbf{Content out of anchor. Incomplete response.}}
\end{casebox}

\subsection{DIA}

\begin{casebox}[colback=green!5]{Case 1: Successful Generation}
    \textbf{[Query]} \\
    Charlie wants to sell beeswax candles. For every pound of beeswax, he can make 10 tapered candles. One pound of beeswax and the wicks cost \$10.00 in supplies. If he sells each candle for \$2.00 each, what is his net profit if he makes and sells 20 candles?
    
    \tcblower
    
    \textbf{[Output]} \\
    \texttt{""<think>First, I need to figure out how much candles Charlie can make with 20 candles... [Reasoning Steps] ... Therefore, Charlie will profit a net of \$20.00 and makes and sells 20 candles0</think><answer> 20</answer>""}
    
    \vspace{0.5em}
    \textbf{[Flaw]} \textcolor{green!40!black}{\textbf{None.}} (High quality reasoning and format compliance).
\end{casebox}

\begin{casebox}{Case 2: Minor Length Prediction Error}
    \textbf{[Query]} \\
    John drives for 3 hours at a speed of 60 mph... (omitted for brevity) ... How far is he from home at the end of those 4 hours?
    
    \tcblower
    
    \textbf{[Output]} \\
    \texttt{""<think>... [Correct Reasoning Steps] ... Since, he traveled 180 miles away from home and then covered 175 miles back,,, he is distance from home at 180 - 175 = 5 miles. the end of those 4 hours, he, he, 55555555555555 miles5</think><answer>5 miles</answer>""}
    
    \vspace{0.5em}
    \textbf{[Flaw]} \textcolor{orange!80!black}{\textbf{Generation length prediction not completely accurate.}} (Model pads the end of reasoning with repetitive tokens to fill predicted length).
\end{casebox}
\section{dLLM Prompt Template}
\label{appendix Prompt Template}
\begin{figure}[h]
    \centering
    \begin{tcolorbox}[
        colback=gray!10,
        colframe=gray!60,
        title=\textbf{Prompt Template for Reasoning Task}, 
        fonttitle=\bfseries\small,
        arc=1mm,
        width=\linewidth,
        boxsep=2pt,
        left=3pt, right=3pt,
        top=2pt, bottom=2pt,
        fontupper=\small\ttfamily,
        boxrule=0.8pt]
        \textbf{[Instruction]} \\
        You are a helpful assistant that helps the user to solve the question.
        
        \vspace{0.3em}
        \textbf{[Output Format]} \\
        You need to think first and then answer the question briefly by following the format: 
        \textless think\textgreater...\textless/think\textgreater
        \textless answer\textgreater...\textless/answer\textgreater.
        
        \vspace{0.3em}
        \textbf{[Input]} \\
        Here are the questions: \{QUESTION\}
    \end{tcolorbox}
    \caption{The prompt template used for guiding the model to generate structured reasoning chains.}
    \label{fig:prompt_design}
\end{figure}

\section{Stage Ablating Experiment on Reasoning Benchmarks}
Table~\ref{tab:dia_ablation_appendix} details our stage ablation. Stage 1 (confidence prediction) is crucial for strict format adherence. Stage 2 (iterative denoising) is indispensable; ablating it prevents generating any concrete responses.
\begin{table*}[t]
    \centering
    \begin{tabular}{l ccc ccc}
        \toprule
        & \multicolumn{3}{c}{\textbf{GSM8K}} & \multicolumn{3}{c}{\textbf{MATH}} \\
        \cmidrule(lr){2-4} \cmidrule(lr){5-7} 
        \textbf{Method} & \textbf{Acc.} & \textbf{$S_{format}$} & \textbf{Latency} & \textbf{Acc.} & \textbf{$S_{format}$} & \textbf{Latency} \\
        \midrule
        \textbf{DIA (Ablated Stage 1)} & 10.31 & 0.00 & 14.99 & 6.73 & 0.84 & 15.33 \\
        \textbf{DIA (Full Method)} & \textbf{47.54} & \textbf{59.67} & 25.86 & \textbf{20.20} & \textbf{75.62} & 29.37 \\
        \bottomrule
    \end{tabular}
    \caption{Detailed results of DIA method ablation (Full Metrics)}
    \label{tab:dia_ablation_appendix}
\end{table*}

\begin{table*}[t]
    \centering
    \begin{tabularx}{\textwidth}{l *{9}{>{\centering\arraybackslash}X}}
        \toprule
        & \multicolumn{9}{c}{\textbf{GSM8K}} \\
        \cmidrule{2-10}
        & \multicolumn{3}{c}{\textbf{$C = 0.035$}} & \multicolumn{3}{c}{\textbf{$C = 0.05$}} & \multicolumn{3}{c}{\textbf{$C = 0.065$}} \\
        \cmidrule(lr){2-4} \cmidrule(lr){5-7} \cmidrule(lr){8-10}
        & \textbf{Acc.} & \textbf{$S_{format}$} & \textbf{Latency} & \textbf{Acc.} & \textbf{$S_{format}$} & \textbf{Latency} & \textbf{Acc.} & \textbf{$S_{format}$} & \textbf{Latency} \\
        \midrule
        \textbf{$\Delta = 2$} & 14.48 & 90.60 & 17.27 & 14.25 & 90.60 & 17.23 & 14.71 & 89.54 & 17.70 \\
        \textbf{$\Delta = 4$} & 47.31 & 60.42 & 25.72 & 47.54 & 59.67 & 25.86 & 47.54 & 58.91 & 26.64 \\
        \textbf{$\Delta = 8$} & 48.60 & 57.32 & 25.78 & 48.37 & 56.71 & 25.98 & 48.37 & 56.56 & 26.52 \\
        
        \midrule
        \midrule
        & \multicolumn{9}{c}{\textbf{MATH}} \\
        \cmidrule{2-10}
        & \multicolumn{3}{c}{\textbf{$C = 0.035$}} & \multicolumn{3}{c}{\textbf{$C = 0.05$}} & \multicolumn{3}{c}{\textbf{$C = 0.065$}} \\
        \cmidrule(lr){2-4} \cmidrule(lr){5-7} \cmidrule(lr){8-10}
        & \textbf{Acc.} & \textbf{$S_{format}$} & \textbf{Latency} & \textbf{Acc.} & \textbf{$S_{format}$} & \textbf{Latency} & \textbf{Acc.} & \textbf{$S_{format}$} & \textbf{Latency} \\
        \midrule
        \textbf{$\Delta = 2$} & 7.40  & 90.38 & 23.52 & 7.58  & 90.42 & 24.81 & 7.82  & 89.60 & 26.02 \\
        \textbf{$\Delta = 4$} & 18.30 & 78.96 & 30.55 & 18.88 & 77.96 & 31.72 & 19.48 & 75.98 & 34.12 \\
        \textbf{$\Delta = 8$} & 20.20 & 75.62 & 29.37 & 20.54 & 75.00 & 30.62 & 21.64 & 69.67 & 30.42 \\
        \bottomrule
    \end{tabularx}
    \caption{DIA results in different hyper parameter combinations.}
    \label{tab:hyperparameters}
\end{table*}
\section{Reasoning Benchmark Results under different Hyper Parameters}
We analyzed parameter sensitivity in Table \ref{tab:hyperparameters}. Threshold $C$ dictates truncation certainty, while $\Delta$ governs expansion aggressiveness. A smaller $\Delta$ causes premature truncation (yielding high format retention), whereas a larger $\Delta$ provides sufficient reasoning space (boosting accuracy). Thus, $\Delta$ acts as a flexible knob balancing format strictness and reasoning depth, achieving optimal performance when paired with confidence-based prediction.

\section{Usage of LLMs}
\label{appendix Usage of LLMs}
We acknowledge the use of large language models (LLMs) in the preparation of this paper. LLMs were employed exclusively as writing assistance tools for language polishing, grammar refinement, and readability improvement. They were not involved in research ideation, experimental design, or data analysis. All technical ideas, theoretical developments, proofs, and experimental results presented in this paper are entirely the work of the authors, who take full responsibility for the accuracy and integrity of the final submission.

\section{Ethics and Transparency Statement}

\textbf{Artifacts and Licensing}: All artifacts used in this study are open-source and utilized for research purposes. The GSM8K and MATH datasets are provided under the MIT License. The WikiBio dataset is distributed under the Creative Commons Attribution-ShareAlike 3.0 Unported License (CC BY-SA 3.0). The baseline Dream-7B models and their associated codebase are licensed under Apache License 2.0. Computational frameworks, including PyTorch, are used under their respective open-source licenses (\textit{e.g.}, BSD-style). 

\textbf{Intended Use Compliance}: We confirm that our use of the GSM8K, MATH, and WikiBio datasets, as well as the Dream-7B models, is strictly for academic research purposes. This is fully consistent with their intended use as benchmarks and base models for evaluating and enhancing the reasoning and generation capabilities of large language models. Specifically, our use of the WikiBio dataset aligns with its original purpose of evaluating text generation algorithms, and we strictly adhere to its attribution and share-alike requirements.

\textbf{Documentation}: We provide a concise description of the artifacts used in this study, covering their target domains, supported languages, and task specifications to ensure a comprehensive understanding of the experimental context:
\begin{itemize}
    \item \textbf{GSM8K:} Consists of high-quality grade school math word problems. It covers multi-step arithmetic reasoning in the \textbf{English} language.
    \item \textbf{MATH:} A challenging benchmark spanning competition-level problems from elementary to advanced mathematics (e.g., Algebra, Calculus) in \textbf{English}.
    \item \textbf{WikiBio:} A dataset comprising 728,321 biographies extracted from the English Wikipedia dump. It provides tokenized infoboxes and the corresponding first paragraphs of articles, primarily serving as a benchmark for evaluating structured data-to-text generation.
    \item \textbf{Dream-7B:} A series of diffusion large language models designed for general-purpose language understanding and instruction following, with specific optimizations for parallel generation.
\end{itemize}

\end{document}